\documentclass[letterpaper,twocolumn,10pt]{article}
\usepackage{usenix-2020-09}
\usepackage{microtype}
\usepackage{graphicx}
\usepackage{subfigure}
\usepackage{booktabs} 
\usepackage{xurl}

\usepackage{hyperref}
\usepackage{xspace}
\usepackage{comment}

\usepackage{multirow}
\usepackage{array}
\usepackage{tabularx}
\usepackage{makecell} 
\usepackage[authoryear,round]{natbib}
\usepackage{authblk}

\newcounter{packednmbr}
\newenvironment{packedenumerate}{\begin{list}{\thepackednmbr.}{\usecounter{packednmbr}\setlength{\itemsep}{0.5pt}\addtolength{\labelwidth}{-4pt}\setlength{\leftmargin}{3ex}\setlength{\listparindent}{\parindent}\setlength{\parsep}{1pt}\setlength{\topsep}{0pt}}}{\end{list}}
\newenvironment{packeditemize}{\begin{list}{$\bullet$}{\setlength{\itemsep}{0.5pt}\addtolength{\labelwidth}{-4pt}\setlength{\leftmargin}{3ex}\setlength{\listparindent}{\parindent}\setlength{\parsep}{1pt}\setlength{\topsep}{0pt}}}{\end{list}}

\newcommand{\tightcaption}[1]{\vspace{-0.27cm}\caption{{\normalfont{\textit{{#1}}}}}\vspace{-0.2cm}}
\newcommand{\tightsection}[1]{\vspace{-0.25cm}\section{#1}\vspace{-0.15cm}}

\newcommand{\eg}{{\it e.g.,}\xspace}
\newcommand{\ie}{{\it i.e.,}\xspace}

\newcommand{\mypara}[1]{\vspace{0.2cm}\noindent{\bf {#1}:}~}

\newcommand{\name}{\textsc{LMCache}\xspace}
\title{LMCache: An Efficient KV Cache Layer for Enterprise-Scale LLM Inference}
\author{
\normalsize
  Yuhan Liu\textsuperscript{12*}
  Jiayi Yao\textsuperscript{12*}
  Yihua Cheng\textsuperscript{1*}
  Yuwei An$^1$
  Xiaokun Chen$^1$
  Shaoting Feng\textsuperscript{12} \\\vspace{-10pt}
  Yuyang Huang\textsuperscript{12}
  Samuel Shen$^1$ 
  Rui Zhang$^1$
  Kuntai Du$^1$
  Junchen Jiang$^1$ \\
  \textsuperscript{1}Tensormesh Inc.
  \textsuperscript{2}University of Chicago
}
\begin{document}
\maketitle
\begingroup
  \renewcommand\thefootnote{} 
  \footnotetext{*Equal contribution} 
\endgroup

\vskip 0.3in

KV cache has traditionally been stored in GPU memory to accelerate the decoding phase of large language model (LLM) inference.
However, it is increasingly necessary to move KV caches outside GPU devices, to enable cache reuse across different queries, and inference engines. Our real-world usage statistics confirm this trend: over time, the total KV cache stored by users has grown rapidly, far exceeding the capacity of GPU memory.
Despite this need, there lacks an efficient solution for offloading and transferring KV caches.
We present {\bf \name}, the first and so far the most efficient open-source KV caching solution, which extracts and stores KV caches generated by modern LLM engines (vLLM and SGLang) out of the GPU memory and shares them across engines and queries.
\name supports both {\em cache offloading} (prefix reuse across queries) and {\em prefill–decode (PD) disaggregation} (cross-engine/GPU cache transfer).
\name's high performance and wide adoption stem from the following contributions:
{\em (i)} highly optimized KV cache data movement powered by batched data movement operations, compute and I/O pipelining;
{\em (ii)} a modular KV cache connector component, decoupling \name from the rapid evolution of inference engines;
{\em (iii)} a first-class control API, such as pinning, lookup, cleanup, movement, and compression, for flexible cache orchestration across GPU, CPU, storage, and network layers.
Our evaluation shows that combining \name with vLLM achieves up to 15$\times$ improvement in throughput across workloads such as multi‑round question answering and document analysis.
Large-scale adoption of \name in enterprise settings provide us valuable insights, for example, fetching KV cache from remote storage has unsurprisingly benefits to prefill delay, and that context truncation, which is a widely applied technique in industry, can greatly reduce prefix cache hit ratio by half.   
The source code of \name is at: \url{https://github.com/LMCache/LMCache}.

\vspace{20pt}

\section{Introduction}

Today, large-language model (LLM) {\em inference} has outpaced training in growth. 
LLM inference powers millions of applications, from interactive customer support and code generation to retrieval-based document analysis and agentic workflows.
To build the systems for LLM inference, 
KV cache, the intermediate states of large language model (LLM) inference, has now become a de facto optimization to make inference faster. 

Traditionally, KV cache has been used during new token generation to skip re-computing the KV cache of the input prompts, \emph{within a single query}. 
Thus, every query is processed independently by \emph{one instance} of the inference engine, and a LLM query's lifecycle, including computation and I/O operations, happens in the GPUs and the GPU memory of one inference engine. 
\begin{figure}[t!]
\centering
    \includegraphics[width=\columnwidth]{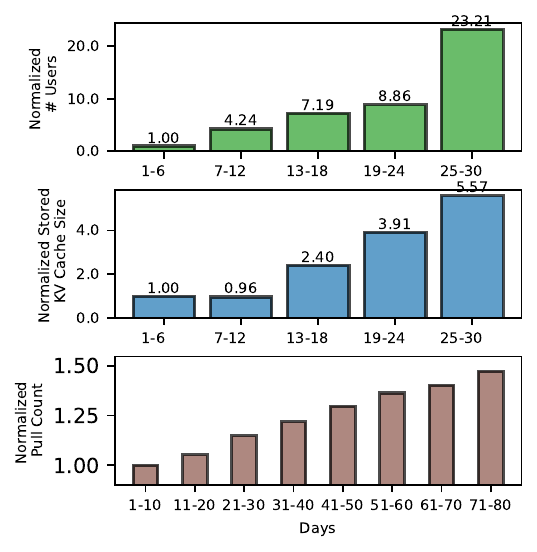}
    \vspace{-10pt}
    \tightcaption{Upper: \name is being used by more and more users over time. Middle: \name is being used to store KV cache of larger size. Bottom: \name's docker image pull counts have continued to increase.   }
    \label{fig:intro_usage_results}
\end{figure}

However, recent trends has advanced to propose moving KV cache outside the GPU memory, including two emerging directions: 

\mypara{Cross-query caching to avoid redundant compute} KV cache can be \emph{persisted} (\eg to lower-tier storage devices) beyond the lifecycle of a query to avoid any re-computation of the shared prefix for another query.

\mypara{Prefill–decode disaggregation for higher utilization} There is an arising trend to decouple the prefill and decode phase on different GPUs, to ensure that latency-sensitive decoding phase is not affected by throughput-oriented prefill phase. Such prefill-decode (PD) diaggregation requires KV cache produced by the prefill GPUs to be transferred to the decode GPUs. 

Our real-world usage statistics confirm the trend of moving KV cache out from the GPU memory. 
As will be discussed in detail soon (\S\ref{subsec:usage}), over time, the total size of KV cache that is stored by the users continues to grow, and it far exceeds the capacity of the GPU memory. 
This suggests that KV cache may need to be frequently evicted from GPU memory as more and more requests come in. 
Also, in order for it to be reused by another query, it needs to be offloaded from GPU memory, and then loaded back to GPU memory for reuse. 

To make these ideas practical, the LLM inference systems must be augmented with new KV cache semantics. 
In particular, inference engines should support the new interface that {\em extracts} KV caches from a normal inference call and {\em re-loads} KV caches into subsequent queries on demand. 
The system also must allow the extracted KV caches to be {\em stored} persistently and {\em transferred} across distributed inference engines. 
Most importantly, for such interface extensions to be practical, KV cache extraction, re-loading, storage, and transfer must be efficient, and the new interfaces must remain compatible with rapidly evolving inference engines such as vLLM~\cite{vllm} and SGLang~\cite{zheng2024sglangefficientexecutionstructured}.

We introduce \textbf{\name}, the first open-source library that provides a high-performance implementation of these new KV cache semantics.
With \name, KV cache can be extracted from and loaded back to inference engines efficiently, stored in a hierarchy of storage devices (CPU memory, local disk, remote disk, and Redis), and transferred over different networks (Ethernet, RDMA, NVLink). 

\name makes three distinct contributions.

\mypara{\#1. Highly optimized performance} 
\name incorporates a series of performance optimizations that make storing and loading KV cache efficient and practical in real deployments. 
For instance, \name batches operations to pipeline the storing and loading of KV cache, as well as to pipeline GPU compute and data loading/storing (\eg loading next layer's KV cache while performing computations for the current layer).
Moreover, rather than storing/loading KV cache at the granularity of the inference engine's native small page size, \name stores/loads KV cache at a configurable chunk size, often much larger than page size, to fully utilize the bandwidth between storage devices and GPU memory. 
\name also minimizes the copies of KV cache data when moving them among different storage tiers, by implementing zero-copy operations. 

\mypara{\#2. Standardized interface with inference engines} 
\name defines standardized connector interfaces that remain compatible with fast-changing inference engine backends. 
On average, 15--20 new open-weight models are released every week in 2025, so to best utilize new hardware for the new models, modern LLM inference engines must evolve rapidly, potentially changing the KV cache layout in GPU memory and thus affecting the \name interface. 
    To address this, \name designs and implements a modular KV cache connector interface that decouples \name with the inference engine backend, so \name can easily adapt to the evolving APIs in the inference engines. 
    
\mypara{\#3. Flexible KV cache management interface} The interface augmentation introduced by \name exposes KV cache, a new data structure in LLM inference. 
    \name exposes APIs that allow developers and operators to locate, move, pin, and even compress KV cache extracted from inference engines.
    These first-class APIs allow higher-level applications, such as query schedulers or routers, to make better decisions, such as KV cache-aware query routing.

Our evaluation demonstrates that \name consistently outperforms both built-in KV caching mechanisms in open-source inference frameworks and commercial inference APIs,  delivering up to 15$\times$ higher throughput and at least 2$\times$ lower latency across diverse settings, including local prefix caching, distributed prefix reuse, and PD disaggregation.

Beyond quantitative gains, \name has seen adoption across several enterprises and open-source projects, providing useful insights and lessons in KV cache-driver optimizations at production scale. 
This includes surprising gain in latency brought by remote storage backend, reduction in prefix cache hit rate because of context truncation, and that the flexibility to evolve faster and integration smoothness is more important than the language performance itself.

The remainder of this paper details the motivation (\S\ref{sec:motivation}) and challenges (\S\ref{sec:challenges}), \name architecture and key design choices (\S\ref{sec:overview}, \S\ref{sec:design}, and \S\ref{sec:connector}), deployment experiences (\S\ref{sec:deployment}) and experimental evaluation (\S\ref{sec:eval}).

\begin{figure*}[t!]
\centering
    \includegraphics[width=.9\linewidth]{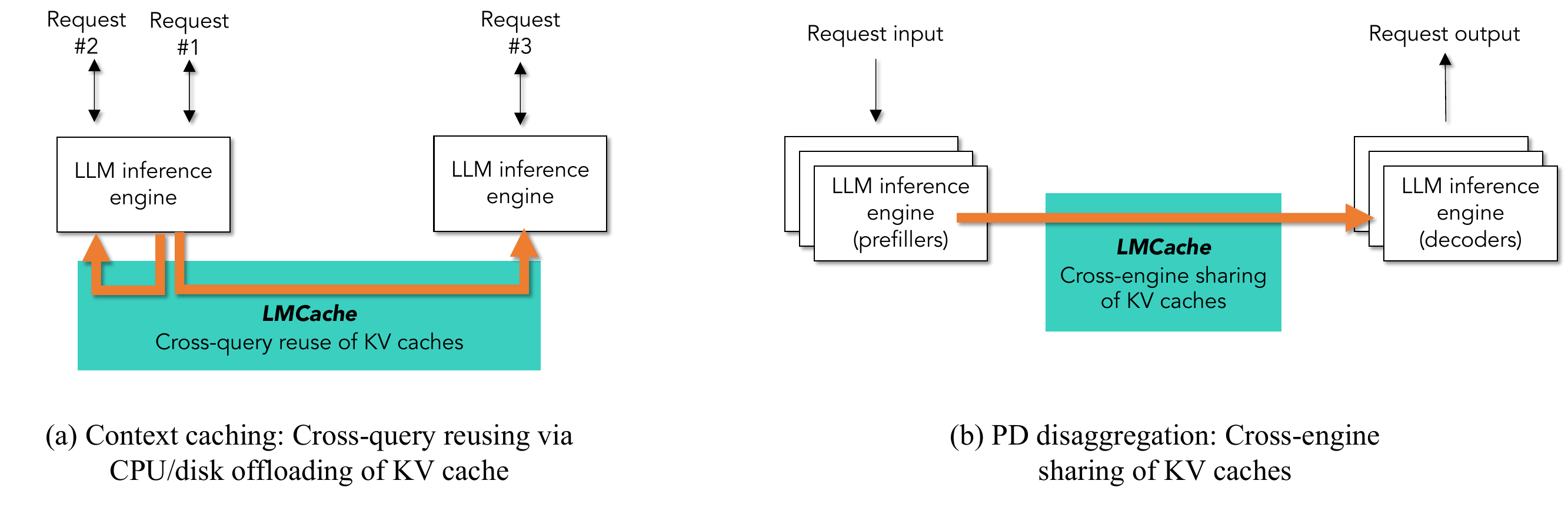}
    \tightcaption{\name supports both context caching (KV cache offloading and sharing across queries) and PD disaggregation (cross-engine transfer of KV caches.}
    \label{fig:usecases}
\end{figure*}

\section{Motivation and Real-world Usage Statistics}
\label{sec:motivation}
\subsection{KV Cache in LLM Inference}


{\em KV cache} was originally introduced to accelerate a single inference query by storing the attention states, in the form of $K$ and $V$ tensors, for input tokens and previously generated tokens directly in GPU memory. 
KV cache effectively stores the attention information between each pair of tokens that have been seen so far in this query. 
In short, it is a {\em LLM-native} representation of knowledge. 

Nowadays, the contexts have grow longer and longer, and people have started to augment inference with background knowledge. 
Given this trend, it is popular to share KV cache across different user queries to reduce the redundant computations for the long contexts or background knowledge.


\begin{figure}[t!]
\centering
    \includegraphics[width=\columnwidth]{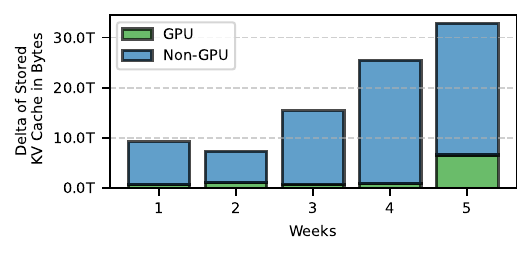}
    \vspace{-20pt}
    \tightcaption{Weekly growth of KV cache size, including portions that fit in GPU memory and those that exceed it.}
    \label{fig:motivate_non_gpu_size}
\end{figure}

\subsection{Real-world Usage Statistics}
\label{subsec:usage}
\mypara{KV Cache Size Exceeds GPU memory} Although KV cache has been kept inside GPU memory for all traditional LLM inference systems, we observe that the required size of KV cache is now far exceeding the GPU memory capacity, from our real-world usage statistics, gathered by usage tracker voluntarily turned on by users. 

Figure~\ref{fig:motivate_non_gpu_size} shows the weekly growth of KV cache size over the past five weeks, for caches that fit within GPU memory (green) and those that exceed GPU memory capacity (blue).
The portion of KV cache that no longer fits in GPU memory has increased significantly over time, showing that GPU memory alone is insufficient for storing all caches. 
To enable KV cache reuse across queries, especially those generated long before reuse, it becomes necessary to move KV cache out of GPU memory, for instance, by offloading it to CPU memory or other storage tiers.

\mypara{Reuse per Token has Greatly Increased} We also observe that reuses per token has greatly increased over time.  

\begin{figure}[t!]
\centering
    \includegraphics[width=\columnwidth]{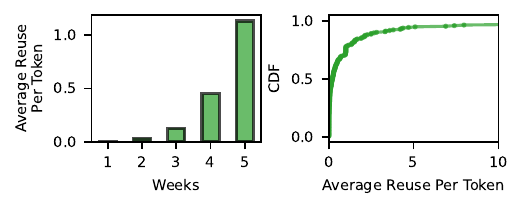}
    \vspace{-20pt}
    \tightcaption{Left: Average reuse per token for top users. Right: Distributions of average reuse per token across different users. }
    \label{fig:average_reuse_per_token}
\end{figure}

As shown on the left side of Figure~\ref{fig:average_reuse_per_token}, where we plot the ratio between reused tokens and all stored tokens, beyond the GPU memory, plotted with top-10 users.  
We denote this ratio as reuses per token. 
The reuses per token has grown significantly over the past several weeks, which indicates that tokens that cannot fit inside GPU memory are being more and more frequently reused by inference.
This suggests that more and more tokens need to be loaded back to GPU memory. 

On the right hand side, the figure shows the distribution of reuse per token for different users over the past week. 
More than 19\% of users reuse stored tokens for more than 1.5 times, suggesting the trend of users accessing a token multiple times after it is stored. 

\subsection{Need an Efficient KV Caching Layer for Moving KV Cache}

From the above observations from the statistics gathered in real-world deployments, we find two important trends of KV caching. 
First, the KV cache that cannot simply fits in GPU memory keeps growing, potentially due to growing length of the contexts or larger amount of user traffic. 
Second, reuses per token stored beyond the GPU memory has also increased over time.
Both trends suggest that we need to move KV cache out of GPU memory. 
Specifically, in the current industry, two scenarios which move KV cache out from GPU exist:

\begin{packedenumerate}
\item {\em Context caching (\ie {\em cross-query} KV cache reuse)} persists KV cache segments from one query and reusing them for subsequent queries that share a common prefix. Examples include document analysis where the same document (chunk) remains constant across multiple queries, and multi-turn dialogues with a fixed system prompt or long preamble.
Prefix caching reduces redundant computation during the prefill phase, directly lowering TTFT and GPU-hours per query~\cite{liu2024cachegenkvcachecompression, gao2024costefficientlargelanguagemodel, cake, 10.1145/3719330.3721230, mooncake, jin2024ragcacheefficientknowledgecaching, chen2024kvdirectdistributeddisaggregatedllm, impress}.

\item {\em Prefill--decode (PD) disaggregation (\ie {\em cross-engine} KV cache transfer)} splits inference into a \emph{prefill} stage (processing the entire input prompt) and a \emph{decode} stage (autoregrssive token generation) across different GPUs or nodes. 
This approach reduces tail latency by maximizing the decoding speed without being interrupted by the prefill phase~\cite{distserve, splitwise, shi2025nexusproactiveintragpudisaggregationprefill}. 

\end{packedenumerate}

However, there lacks a library to support efficient extraction and loading from and to the GPU memory due to the system challenges as discussed soon (\ref{sec:challenges}).

\section{Challenges of Efficient KV Caching and Related Work}
\label{sec:challenges}
\subsection{Challenges of Efficient KV Caching}
Despite their potential, the practical adoption of prefix caching and PD disaggregation is limited by three interrelated systems challenges:

\begin{table}[]
\centering
\begin{tabular}{@{}cc@{}}
\toprule
Message Size & Transfer Throughput \\ \midrule
64KB         & 4GBps               \\
256KB        & 13GBps              \\
1MB          & 30GBps              \\
10MB         & 46GBps              \\
16MB         & 49GBps              \\
100MB        & 49GBps              \\ \bottomrule
\end{tabular}
\tightcaption{Transfer message size vs achieved transfer throughput using RCCL transfer library~\cite{UCCL2025KVTransferEngine}. }
\label{tab:size_throughput}
\end{table}

\subsubsection{Challenge \#1: I/O inefficiency under paged memory} 

KV cache storage and transfer used to rely on PyTorch serialization (\texttt{torch.save} / \texttt{torch.load}) or primitive tensor copying, with a typical transfer speed of only sub-1GB/s.  
These methods introduce non-trivial delay overhead, especially when handling large data structures like KV caches, and lack zero-copy support with various storage devices (local or remote), causing extra CPU-GPU data copies.

Recent high-throughput inference engines, such as vLLM~\cite{vllm} and SGLang~\cite{zheng2024sglangefficientexecutionstructured}, make KV cache storage and transfer even more challenging. 
They employ \emph{paged} attention memory, dividing the attention buffer into small, fixed-size pages (typically 16--64 KB). 
For instance, vLLM uses 62.5-KB page in Llama-3.1-8B-Instruct model. 
The paged memory architecture is widely used because it improves batching and memory utilization.

However, because the pages of a KV cache are not always contiguous, the paged memory architecture dramatically increases the number of small-sized I/O operations required to persist or transfer a KV cache. 
Transferring such small chunks of data is known to suffer from network bandwidth {\em under}utilization and reduce throughput~\cite{kwon2023demystifyingnccl, nvidia2020smalltransfer, facebook2024roce}. 
Prior work (Table~\ref{tab:size_throughput}) has shown that, on a setup with two AMD GPU nodes connected by eight Broadcom Thor-2 400Gbps NICs, the transfer size must reach at least 16 MB to saturate the available network bandwidth~\cite{uccl_transport}.
Furthermore, prior work has shown that only transferring a data size of megabyte range (\eg 1--2MB) can achieve 75--80\% of the theoretical PCIe 5.0 bandwidth~\cite{strata}. 



\subsubsection{Challenge \#2: Compatible with fast-evolving inference engines} 

With the widespread use of AI, new LLMs and hardware accelerators are introduced at a rapid pace. 
In 2025, one prominent LLM was released on average every 4 days~\cite{best44llms2025}.
In response, inference engines must evolve just as quickly. 

Each update to accommodate new models or hardware often changes GPU memory allocation, which in turn changes the KV cache interface. 
For example, when vLLM adopts a new attention kernel that produces KV caches with different dimensions, the KV caching library must be updated to translate the new kernel’s output KV cache format into one compatible with the KV cache library. 
Keeping up with these frequent changes requires tremendous effort, given the fast-moving inference engines.




\subsubsection{Challenge \#3: Lack of management APIs} 

As KV caching becomes a first-class citizen in the LLM inference backend, various components (in addition to the LLM inference engines), as well as ML ops teams, will need to make decisions in a KV-cache-aware manner. 
Yet, without a unified management interface to locate, evict, pin, or compress caches, these upper-layer modules cannot make informed placement or eviction decisions. 
This leads to inefficient cache utilization, duplicated storage, and unpredictable eviction policies.
For instance, inference query routers, which assign each query to one of the inference engine instances, need to know the locations of KV caches, in order to route queries to instances that already hold the KV cache for matched prefix tokens locally (\eg in CPU memory).

Moreover, applications now also demand such KV-cache management interfaces. 
In early 2025, for instance, a financial company\footnote{For confidentiality, we do not disclose names of enterprise users in this report.} that has worked closely with \name in the production setting asked for an interface that allows users to {\em explicitly} pin frequently accessed financial documents in the KV caching system, for more efficient access to popular contexts.
As another example, an agent company requested a series of APIs that allow them to identify the KV cache of a given content, compress the KV cache, and transfer the compressed KV cache across nodes.


\subsection{Related Work and Existing Solutions}

Several KV cache handling mechanisms exist, but none of them fully address the above challenges:

\mypara{Inference frameworks} Since the release of vLLM Production Stack~\cite{vllm-production-stack2025} in January 2025, there have been several open-source distributed inference stacks, including Nvidia's Dynamo~\cite{aiDynamoDynamo2025}, AIBrix~\cite{theaibrixteam2025aibrixscalablecosteffectivelarge}, \texttt{llm-d}~\cite{llmDllmD2025}, SGLang OME~\cite{ome}, and KServe~\cite{KServe2025}. 
They focus on easy deployment of inference engine solutions over Kubernetes, and technically, they all support various query routers based on load or prefix cache awareness and support KV caching, where \name is used in vLLM production stack, Dynamo, llm-d, and KServe. 

\mypara{Inference engine-native KV caching} 
Open-source inference engines, like vLLM and SGLang, also offer native GPU-to-CPU KV cache transfers, but it is designed for single-node inference, so they lack cross-node transfer optimization or hierarchical storage support for KV cache. 
We will evaluate their performance and compare it against \name in \S\ref{sec:eval}.

\mypara{KV cache storage layers}
Mooncake~\cite{qin2025mooncakekvcachecentricdisaggregatedarchitecture}, Redis~\cite{redis-enterprise-software-references}, InfiniStore~\cite{bytedanceInfiniStore2025}, and 3FS~\cite{DeepSeek2025ThreeFS} provide distributed object storage or caching, but they lack an efficient ``glue'' layer between the inference engines to efficiently move small tensors frequently across different storage tiers, or are tightly coupled with a specific inference framework. 

\mypara{Proprietary implementations} Proprietary inference APIs (e.g., Fireworks AI, Together AI) implement their own prefix caching internally, but these are tied to their \emph{closed-source} serving stacks and are not accessible to operators deploying their own infrastructure.

\mypara{Source code for research} 
Several research proposals have open-sourced prototypes for their KV cache optimizations, including prefix caching~\cite{vllm, zheng2024sglangefficientexecutionstructured, pensieve, prompt-cache, chunkattention, infinigen, blendserve, jin2024ragcacheefficientknowledgecaching, gao2024costefficientlargelanguagemodel, impress, cake, kvshare}, PD disaggregation~\cite{distserve, splitwise, shi2025nexusproactiveintragpudisaggregationprefill}, and KV cache compression~\cite{liu2024kivi, kvpress2024, xiao2024duoattentionefficientlongcontextllm, xiao2024efficientstreaminglanguagemodels, snapkv, qin2025cakecascadingadaptivekv, quest, ge2024modeltellsdiscardadaptive, commvq, bitdecoding, pqcache}.
However, these prototypes are typically built on research-oriented inference framworks, such as HuggingFace Transformers, not fully enterprise-ready, or are not designed to evolve alongside the rapidly changing inference engine ecosystem, such as SGLang and vLLM.

\section{Overview of \name}
\label{sec:overview}

\textbf{\name} addresses these challenges by a unified, high-performance KV caching layer capable of efficient storage, movement, and explicit management of KV caches for paged-memory inference engines, making prefix caching and PD disaggregation practical at enterprise scale.

\begin{figure}[t!]
\centering
    \includegraphics[width=.46\columnwidth]{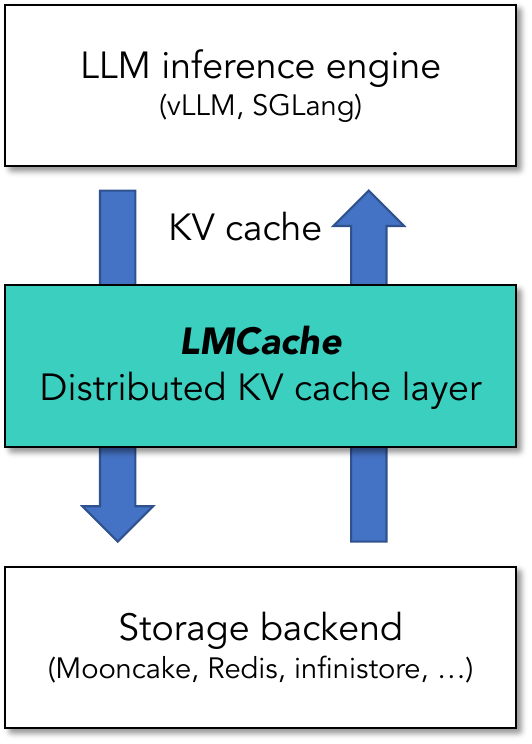}
    \tightcaption{\name sits between LLM inference engines and heterogeneous storage/network devices.}
    \label{fig:high-level}
\end{figure}

As a KV caching layer, \name sits between LLM inference engines and heterogeneous storage/network devices (Figure~\ref{fig:high-level}). 
Its goal is to provide a standardized, high-performance substrate for KV cache movement and management, while remaining compatible with rapidly evolving inference frameworks such as vLLM and SGLang.

\begin{figure*}[t!]
\centering
    \includegraphics[width=0.95\textwidth]{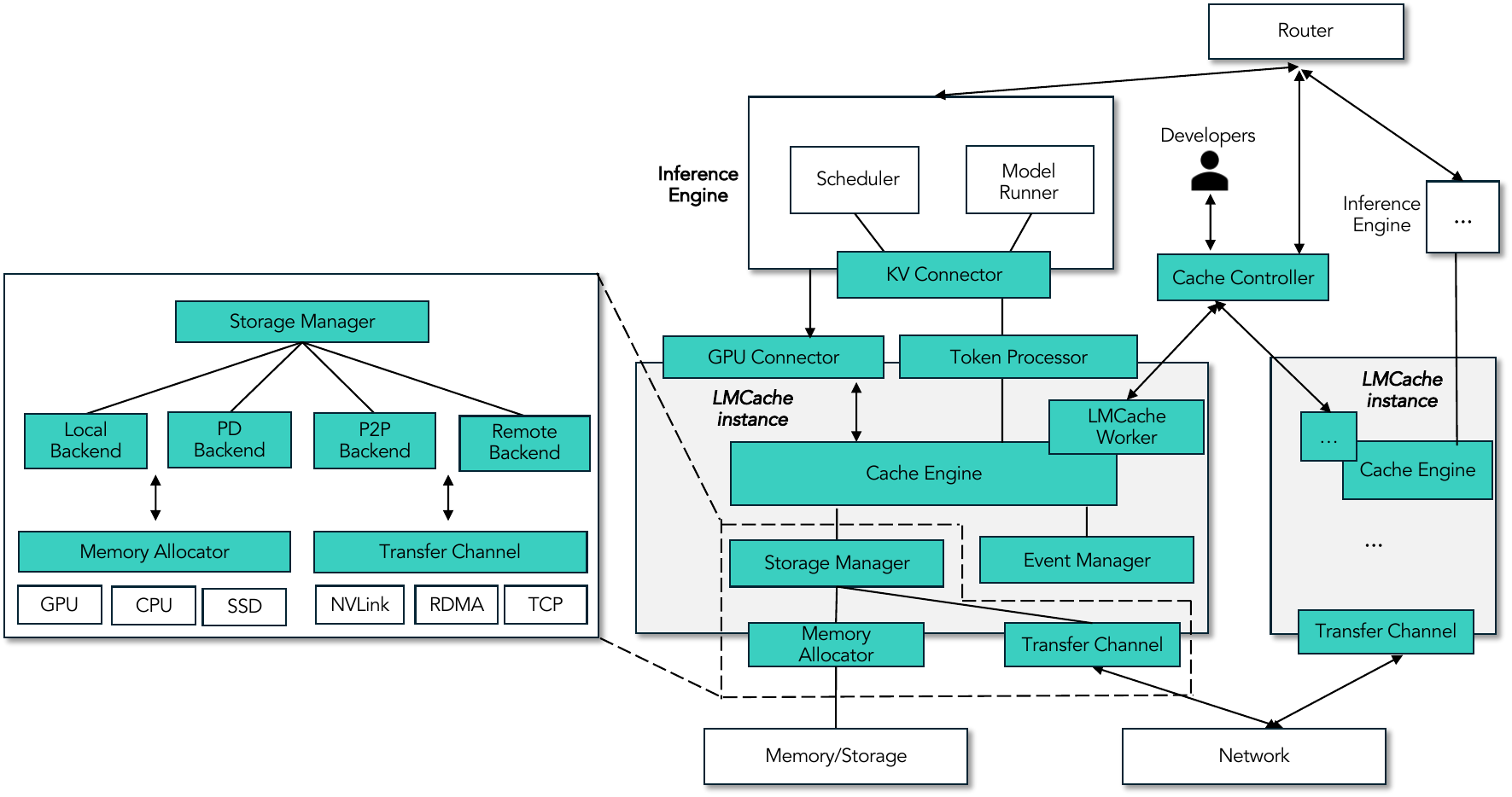}
    \tightcaption{End-to-end system workflow for \name. } 
    \label{fig:architecture}
\end{figure*}

 Figure~\ref{fig:architecture} shows the end-to-end system. Below, we walk through two example workflows: storing and retrieving KV cache.

\mypara{Store} When a new query arrives, it first passes through the \emph{KV connector}, which prepares metadata such as the tokenized input prompt and GPU memory addresses of the relevant pages. The query then goes to the \emph{token processor}, which determines how many new tokens are not yet in the backend and need to be stored. Finally, the storage manager saves the KV cache for these new tokens to the backend via the \emph{transfer channel}, which handles the data transfer logic.

\mypara{Retrieve} When a query requires loading KV cache from the backend, it also starts with the KV connector to prepare metadata. 
The token processor identifies the number of prefix-matched tokens already in the backend. 
Next, the event manager checks if the same query ID has been seen before. 
If so, the cached memory addresses are already tracked and can be returned directly to the \emph{GPU connector}, which loads the KV cache back into GPU memory. 
The event manager also launches asynchronous, layer-wise loading events as described in \S\ref{subsec:io_overlap}. 
If the query ID is new, it is forwarded to the storage manager to look up the CPU memory addresses of the stored KV cache.

\mypara{Lookup} When a query needs to check whether the KV cache for specific tokens exists in the backend, higher-level components such as routers query the cache controller. The cache controller maintains a token pool that records all tokens currently stored in the KV cache backend. 
Whenever a \name instance stores or evicts a KV cache, the \name worker inside the instance updates the token pool with the new status. 
This ensures the token pool always has the up-to-date information of tokens in the backend.




\section{Performance Optimizations}
\label{sec:design}

An important aspect of \name is improving the efficiency of KV cache movement across devices. 
In enterprise-scale LLM inference, \name addresses three key challenges:
\begin{packeditemize}
        \item Modern LLM inference engines manage KV cache at the granularity of pages\footnote{Each page is 16 tokens for a single layer in vLLM. }, 
    which are typically 20~KB--63~KB for popular models including Llama, Qwen, GPT-OSS etc. 
    Such small units are inefficient for transferring, as they cannot saturate bandwidth~\cite{strata, uccl_transport}.
    
    \item KV cache transfers often need to run concurrently with LLM inference. This introduces overhead from two sources. First, data movement can stall inference if transfers are executed in the same CUDA stream as computation. Second, launching memory-copy CUDA functions incurs CPU overhead, as each call consumes CPU cycles and the consumption can be substantial when there are many layers and pages.
    \item During LLM inference, large volumes of queries generate significant amount of KV caches. Duplicating them on any storage device wastes space and introduces copy overhead, which slows down inference.

\end{packeditemize}

Each of these challenges arose from hard lessons in both open-source and enterprise deployments. 
This section describes these challenges in detail and motivates \name's design decisions.

\subsection{Batched Operations}

To address the I/O inefficiency caused by small KV cache units, \name introduces a set of optimizations.

\mypara{Configurable Chunk Size}
Rather than transferring KV cache at the page level, \name groups multiple pages from multiple layers into larger chunks, with a default size of 256 tokens per chunk\footnote{The chunk size is configurable to different I/O speeds. }. 
This is achieved using an intermediate \emph{streaming GPU buffer}. For storing, the KV cache are first copied from the scattered paged GPU memory into a contiguous streaming buffer with a customized CUDA kernel, then offloaded collectively to lower-tier storage (e.g., CPU memory) with DMA engines at the granularity of chunks rather than individual pages. 
For loading, chunks are first retrieved from the storage layer into the GPU buffer with DMA engines and subsequently split into paged memory with CUDA kernels.

\mypara{Parallel Store/load Operations} 
\name supports parallel storage and retrieval of KV caches across multiple storage tiers, including local CPU DRAM or disks, remote CPU DRAM or disks, and object storage (e.g., S3).
In practical LLM serving workloads, KV caches often need to be migrated across devices concurrently—for instance, transferring the KV cache of a hot context from GPU to CPU memory while simultaneously offloading a cold context from CPU memory to local disk.
To maximize link utilization, \name’s store and load APIs accept multiple source and destination devices, enabling concurrent data movement across heterogeneous links. Moreover, these operations can be executed in parallel when the interconnect supports full-duplex communication (e.g., PCIe).



\mypara{Delayed Decode KV Cache Storing}
\name also supports storing newly generated KV caches during decoding.
Instead of offloading each token’s KV cache immediately—a naive approach that triggers frequent small writes—\name buffers KV caches and performs batched storage once a predefined number of tokens (i.e., a chunk) have been generated.
This chunk-based delayed storing strategy reduces write frequency, minimizes I/O overhead, and significantly improves overall storage throughput.

\subsection{Compute-I/O Overlapping}
\label{subsec:io_overlap}

\name employs multiple optimizations aiming for overlapping LLM inference computations with I/O to maximize GPU utilization. 

\mypara{Layer-wise pipelining} \name overlaps KV cache transfers with inference computation through layer-wise pipelining. 
Specifically, it assigns separate CUDA streams for inference computation and data movement within each layer. 
For example, before performing inference on the first layer, its KV cache is loaded into the GPU buffer and transformed into pages. 
While the first layer is running inference, the KV cache for the second layer is asynchronously fetched into the buffer and similarly transformed. 
Note that the second layer's KV cache loading is happened after the first layer's KV cache is put into the right paged memory.
This design ensures that only a fixed-size GPU buffer—whose size is a single layer’s KV cache—is required, while enabling overlapping between data transfer and computation.

\mypara{Asynchronous compute \& prefetch} In many scenarios, there is a time gap between when the inference scheduler admits a query and when the query’s KV cache is actually needed for inference. 
For example, if a query (with cache hit) arrives when the inference engine is processing other queries, the arriving query has to wait in queue.
\name exploits this idle interval to prefetch the queued queries’ KV cache from slower storage tiers into faster ones (\eg from remote disk to local CPU memory or GPU memory). 
As a result, when the actual inference computation starts, the required KV cache can be loaded or used directly from faster-tier storage, significantly reducing loading delay. 
\name allows users to configure the target tier for prefetching based on their own needs in latency SLO and resource constraints.


\subsection{Minimum Data Copy}
A naive implementation of KV cache movement would create additional copies of data at each transfer step, especially when dealing with heterogeneous storage types, leading to redundant memory usage and unnecessary overhead. 
\name avoids this by maintaining only the minimum required copies.

\mypara{Zero-Copy Operations}
When transferring KV cache to multiple devices simultaneously, \name minimizes data duplication through a reference counter. 
Specifically, when KV cache is written to multiple destinations—such as writing from local CPU memory to local disk and remote object storage at the same time—\name increments a reference counter on the shared data for each transfer instead of creating new copies. Each completed read or write decrements the counter, and once the count reaches zero, the data is released. 
This design ensures that data is shared across concurrent read and write operations without unnecessary replication, thus reducing memory pressure and improving efficiency.
This technique is simlar to PCB counter in operating systems~\cite{Strecker1978VAX}. 

\begin{figure}[t!]
\centering
    \includegraphics[width=\columnwidth]{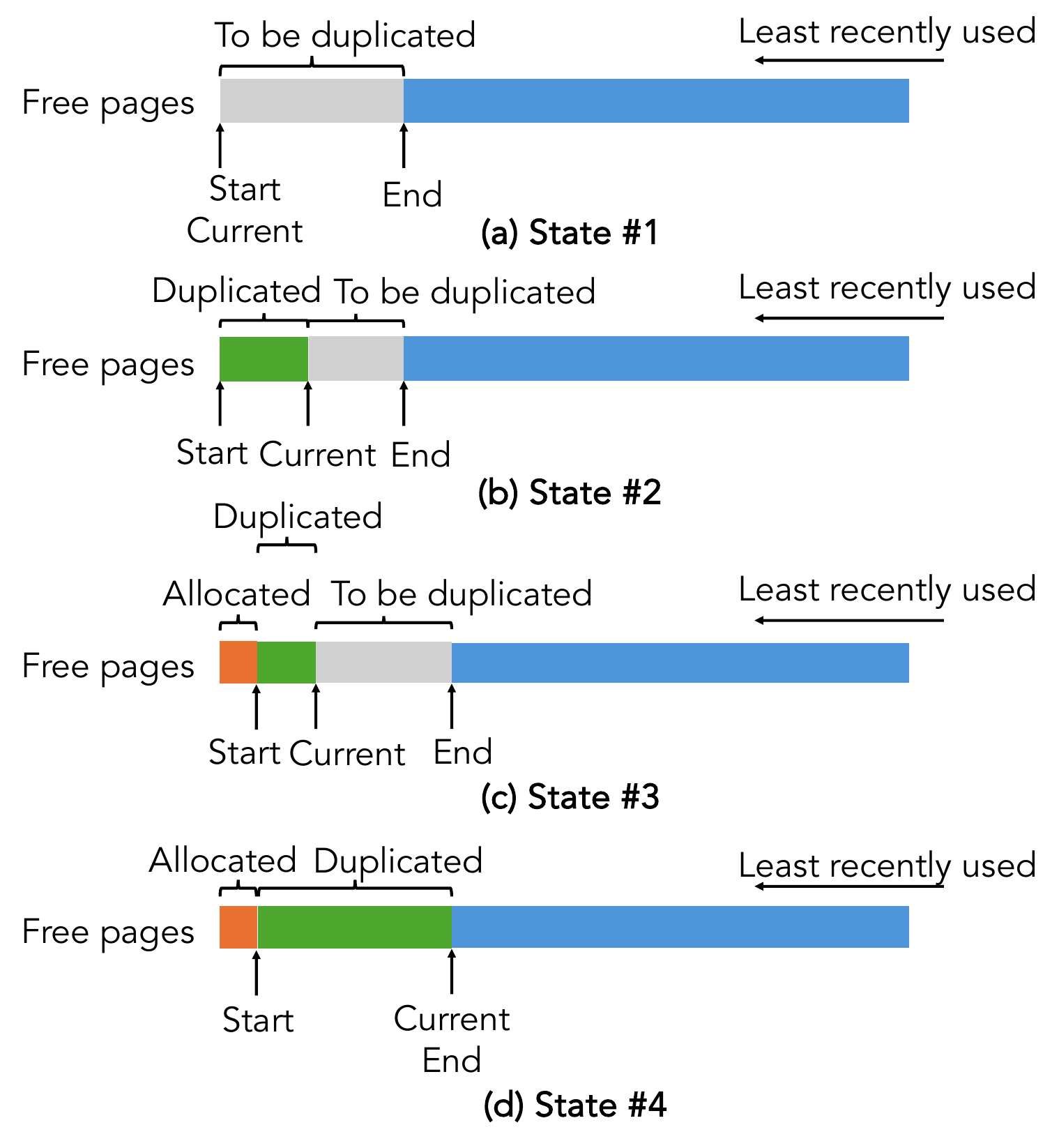}
    \vspace{-15pt}
    \tightcaption{Illustration of dynamic offloading in \name. }
    \label{fig:offloading}
\end{figure}

\mypara{Dynamic Offloading}
Modern inference engines such as vLLM maintain a pool of \emph{free pages} in GPU memory, \ie, pages whose KV cache is not currently used by active queries. 
Instead of duplicating all free pages to CPU memory, \name duplicates only a subset. This mechanism is implemented using three pointers:

\begin{packeditemize}
\item \textbf{Start pointer}: the start address of the free-page region in GPU memory.
\item \textbf{Current pointer}: the index of the free pages that have already been offloaded to CPU memory.
\item \textbf{End pointer}: the end address of the free pages that are scheduled to be offloaded.
\end{packeditemize}

\begin{table*}[t]
\small
\centering
\begin{tabularx}{\textwidth}{l|X}
\hline
Function name & Description \\ \hline
get\_num\_new\_matched\_tokens(query) $\rightarrow$ Optional[matched\_tokens]
& Returns the number of cache-hit tokens found in \name’s backend. Returns None if the \name decides to let vLLM process other requests first and put this request back to waiting queue. \\ \hline
update\_state\_after\_alloc(query, blocks, num\_external\_blocks)
& Updates whether a query needs to transfer KV cache from LMCache's backend. \\ \hline
build\_connector\_meta(scheduler\_output) $\rightarrow$ kv\_connector\_metadata
& Builds metadata for KV cache transfers between LMCache's backend and GPU memory, including GPU memory addresses for KV cache pages. \\ \hline
start\_load\_kv(kv\_pointers)
& Starts loading KV cache from lower-tier storage into GPU memory before LLM inference begins. \\ \hline
wait\_load\_kv(kv\_pointers, 
layer\_id) 
& Synchronizes on KV cache loading to ensure data is available when computation requires it. \\ \hline
start\_store\_kv(kv\_pointer)
& Starts offloading KV cache to lower-tier storage after computation. \\ \hline
wait\_store\_kv(kv\_pointer, 
layer\_id) 
& Synchronizes on KV cache storing to ensure the KV cache for the current layer is offloaded. \\ \hline
\end{tabularx}
\caption{Functions in \name's connector.}
\label{tab:all_func}
\end{table*}

As illustrated in Figure~\ref{fig:offloading}, dynamic offloading has four possible states:

\begin{packeditemize}
\item \textbf{State \#1 (Initialization):} the start and current pointers overlap. The region between the start/current pointers and the end pointer marks the pages pending duplication.
\item \textbf{State \#2 (In-progress):} the current pointer moves toward the end pointer. Pages between the start and current pointers have already been offloaded to CPU memory.
\item \textbf{State \#3 (Query Arrival):} when new queries acquire some of the free pages, the end pointer is moved forward by the number of allocated pages. This ensures sufficient GPU memory is available for future active queries that need to acquire free pages.
\item \textbf{State \#4 (Steady state):} the current pointer overlaps with the end pointer, indicating that all scheduled pages have been duplicated.
\end{packeditemize}

Note that if a query attempts to allocate pages beyond the current pointer, the allocation must stall until the current pointer moves right enough to cover the required pages. 
Thus, a key trade-off in this design is that: the number of duplicated pages—\ie, the region defined by end pointer – start pointer --between GPU and CPU memory. 
A smaller duplication window reduces the duplication ratio but increases the likelihood of allocation stalls. 
For instance, if only one page is duplicated and an inference query requires three pages, the query must wait until the current and end pointers advance by two additional pages. 
On the other hand, if three pages are duplicated, the same query can proceed immediately without stalling, though at the cost of higher duplication ratio. Though not supported, the same dynamic offloading strategies can also be extended to other storage tiers beyong CPU and GPU.



\section{Standardized Interface for Connecting the KV Caching Layer and Inference Engine}
\label{sec:connector}
Modern LLM inference engines, such as vLLM and SGLang, evolve rapidly to support newly released models with diverse architectures. 
For example, in 2025, an average of 15--20 new models are released each week.
Supporting these new architectures often requires non-trivial modifications to inference engines, such as adding support for Sliding Window Attention or Multi-Head Latent Attention. 
These code changes frequently alter how KV cache is managed internally, making it infeasible for \name to adapt in an ad-hoc manner.

To address this challenge, \name introduces a standardized \emph{KV cache connector} interface that decouples KV cache management from the inference engine backend. 
This design ensures that \name remains compatible regardless of how the upstream inference engine evolves.

We note that the design of this API is initiated by LMCache team, but the implementation and the maintainance of this API are the collaborative effort of both LMCache team and vLLM team.

\mypara{Design objectives} The key design objectives are:
\begin{packeditemize}
    \item Maximum flexibility: it enables as much KV cache operations as possible. 
    \item vLLM-native: it aligns with the design direction of vLLM, including strict scheduler --- worker separation, prefix caching as the first-class citizen, and piece-wise CUDA graph, where vLLM only captures CUDA graph for non-attention operations. 
    \item Friendly to out-of-tree connector: it allow intergrating with out-of-tree connectors without vLLM-side code motification.
    \item Minimum API-level overhead: it does not introduce overheads (\eg inter-process communication) at API level.
\end{packeditemize}

To be compatible with the scheduler --- model runner separation philosophy in vLLM, the connector API contains two sets of interfaces:
1) the \emph{scheduler}, where the extra cache-hit tokens from the connector are treated as normal prefix-cached tokens in vLLM and directly influences scheduling decisions and are changed by \name (\ie if there is cache hit in \name, the number of tokens that need to be newly prefilled changes); and  
2) the \emph{model runner}, where we add hooks before and after model execution, and also before and after attention computation, to enable both bulky KV cache offloading and layer-wise KV cache offloading.  





The remainder of this section lists all the interfaces in Table~\ref{tab:all_func}, discusses the design for important APIs, and then traces how a query interacts with these interfaces end-to-end.

The interfaces listed in Table~\ref{tab:all_func} form the foundation of \name’s KV cache loading and storage across lower-tier storage.
Among them, the first three interfaces are implemented within the vLLM scheduler, where they prepare the necessary metadata based on the number of matched tokens found in \name’s KV cache backend.
The remaining four interfaces reside in the model runner, which is responsible for executing the actual KV cache transfers between the inference engine and \name’s KV cache backend.

Putting it together, when a query comes in, the scheduler first calls \texttt{get\_num\_new\_matched\_tokens} which queries \name to see cache hit tokens in the backend. The function can return \texttt{None} if LMCache decides to let vLLM put the current request back to waiting queue and process other requests first, overlapping this request's I/O with other requests' computation.
Then the \texttt{update\_state\_after\_alloc} function decides whether each page in vLLM needs to be loaded from external storage backend based on  matched tokens information from \name. 
If the cache hit tokens are greater than zero, \texttt{build\_connector\_meta} function is called to prepare necessary metadata to load or store KV cache from storage devices. 

Once the query reaches the model runner, in the case of layerwise pipelining, 
\texttt{start\_load\_kv} is called to start loading KV cache of the first layer to GPU memory. 
    Then before each layer's LLM inference computation starts, \texttt{wait\_load\_kv} is called to synchronize the KV cache loading for this layer, and starts the KV cache loading for the next layer. 
 After each layer's inference computation, in the layerwise case, \texttt{wait\_store\_kv} is called to wait until the KV cache for the previous layer has finished storing, and then calls \texttt{start\_store\_kv} to start the storing of KV cache for the newly generated KV cache layer. 

In the case of non-layerwise pipelining, before the first layer's LLM inference starts, \texttt{start\_load\_kv} is called to load the entire KV cache to GPU memory in a blocking manner. 
LLM inference will happen after the KV cache is put to the right GPU memory paged addresses. 
    Then after the LLM inference has done for the current scheduling iteration,  \texttt{start\_store\_kv} is called store the generated KV cache to lower-tier storage synchronously.

\mypara{Impact}
This API is out for over six months in vLLM. During these six months, we see open-source adoptions, including NVIDIA dynamo project, llm-d project from RedHat, AIBrix project from ByteDance, and vLLM production stack project. We also see multiple proprietary connectors from different companies that use the KV connector API.


\begin{table*}[t]
\small
\centering
\begin{tabularx}{\textwidth}{l|X}
\hline
\textbf{Internal APIs}                                                   & \textbf{Description}                                                                                                                                                                          \\ \hline
batched\_admit/batched\_evict(hashes, inst\_id, device) & Send the KV admission/eviction messages from an \name instance to the controller manager. \\ \hline

batched\_p2p\_lookup(hashes) $\rightarrow$ list[inst\_id, device, hit\_chunks]  & Lookup peer KV cache existence from an \name instance based on the the given \texttt{hashes}. \\ \hline

\textbf{External APIs}                                                   & \textbf{Description}                                                                                                                                                                         \\ \hline
lookup(tokens) $\rightarrow$ list[inst\_id, device, hit\_tokens]       & Lookup the global KV cache existence of the given \texttt{tokens}.                                                                                  \\ \hline

move((src\_inst\_id, src\_device), (dst\_inst\_id, dst\_device), tokens)                                & Moves the KV cache of the given \texttt{tokens} from source location \texttt{(src\_inst\_id, src\_device)} to destination location \texttt{(dst\_inst\_id, dst\_device)}.                                                                                                   \\ \hline
clear(tokens, inst\_id, device)                     & Clears the KV cache for corresponding \texttt{tokens} from the storage device \texttt{device} in instance \texttt{inst\_id}.                                              \\ \hline
pin/unpin(tokens, instance, storage\_device)                           & Pins/unpins the KV cache for corresponding \texttt{tokens} at location (\texttt{inst\_id}, \texttt{device}).                                                  \\ \hline
compress/decompress(tokens, instance, device, method) & Compresses/decompresses the KV cache for the corresponding \texttt{tokens} at location \texttt{(inst\_id, device)} with a specified compression/decompression method. \\ \hline
\end{tabularx}
\caption{APIs in \name Controller.}
\label{tab:controller}
\end{table*}

\section{Controller Interfaces}

\name operates as a distributed caching system built around a centralized KV cache controller responsible for global metadata management, cache manipulation, and request routing.
To support these functionalities, \name provides two categories of APIs:
(1) external APIs, which are directly accessible to users or system operators, and
(2) internal APIs, which are used by individual \name instances.

Mechanically, the KV cache controller consists of two layers: a centralized controller manager and per-instance workers. The controller manager runs as a standalone process and serves as a global coordination point, while per-instance workers collocate with each peer \name instance and handle local operations or issue global requests to the manager. External API calls are handled by the centralized manager, which, if necessary, dispatches the appropriate operations to each worker. Per-instance workers can also proactively interact with the centrailzed manager via internal APIs for metadata update or lookup.

The KV cache controller underpins a series of advanced optimizations, including cross-node KV cache sharing, cache-aware request routing, and dynamic KV cache migration. The remainder of this section demonstrates how these optimizations could leverage the controller interfaces through concrete examples.

\mypara{KV cache–aware routing} In this case, higher-level routers aim to direct requests to the instance with the highest expected cache hit rate. Each \name instance reports its cache admission and eviction decisions to the controller manager via the \texttt{batched\_admit} and \texttt{batched\_evict} interfaces. The controller manager aggregates these updates and maintains a global in-memory view of the KV cache state across all instances. When the router calls \texttt{lookup(tokens)}, the controller consults its in-memory global KV cache states and returns a list of \texttt{(instance\_id, storage\_device, hit\_tokens)}, indicating where and how many of the requested tokens are currently cached.

\mypara{KV cache migration} When an instance holding KV cache is about to be scaled down or load balancing is required, the KV cache may need to be migrated to another instance. The controller manager hadles such operations through \texttt{move((src\_inst\_id, src\_deivce), (dst\_inst\_id, dst\_deivce), tokens)} API call by dispatching the request to the source instance. The source instance will try to establish connection to the destination instance if one does not exist, and transfers the specified KV cache from the source storage device \texttt{src\_device} to the destination location indicated by \texttt{(dst\_inst\_id, dst\_deivce)}.  

\mypara{P2P KV cache sharing} \name supports peer-to-peer KV cache sharing, allowing an instance to fetch KV cache from another peer when a local cache miss occurs. Upon a cache miss, the instance's local worker can query the centralized controller manager via \texttt{batched\_p2p\_lookup}. The manager will return a list of \texttt{(inst\_id, device, hit\_chunks)}, representing the number of hit chunks and the location that hold these chunks. The instance can then choose to, for example, load KV cache from a peer with maximum \texttt{hit\_chunks}.

\mypara{KV cache clearance} Applications may clear cache when switching models or reclaiming memory. Upon recieving a \texttt{clear(tokens, inst\_id, location)} call, the controller manager dispatches the operation to the corresponding instance identified by \texttt{inst\_id}. The instance's worker then removes the KV cache associated with \texttt{tokens} stored from a specific stroage device \texttt{device}.  


Some APIs are not covered in the above applications, such as \texttt{compress/decompress(tokens, inst\_id, device, compression\_method)} which compresses/decompresses KV cache stored in location (\texttt{inst\_id}, \texttt{device}) with a specified \texttt{compression\_method} and \texttt{pin/unpin(tokens, inst\_id, device)} which can pin/unpin the specifed KV cache at a certain location \texttt{(inst\_id, device)}.
Users can freely call these APIs to explicitly manage the KV cache in their own applications based on their needs.

\section{Evaluation}


\label{sec:eval}

\begin{figure*}[t]
	\centering
	\includegraphics[]{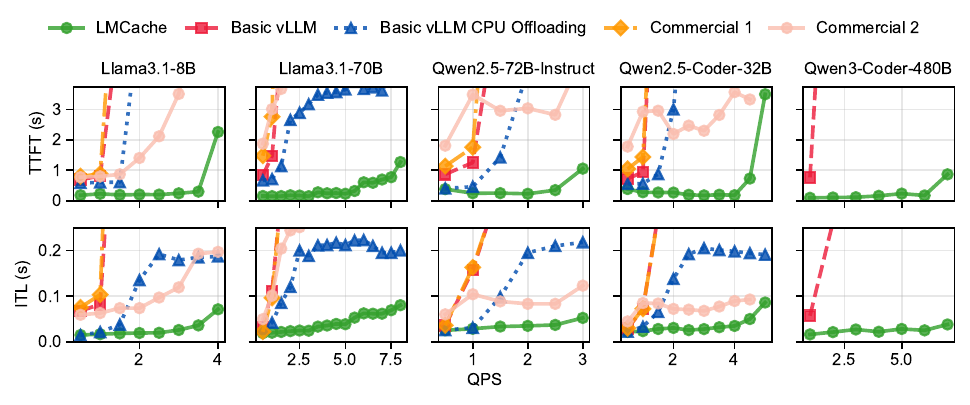}
	\tightcaption{Compared to basic vLLM, basic vLLM CPU offloading, and two commercial alternatives, \name has 1.9--8.1$\times$ smaller TTFT, and supports 2.3--14$\times$ higher inference throughput.  
    Basic vLLM CPU offloading fails to run on \texttt{Qwen3-Coder-480B}, and commercial alternatives do not have the option to deploy \texttt{Qwen3-Coder-480B}. }
	\label{fig:single-node}
\end{figure*}

\subsection{Setup}


\begin{table}[t]
	\centering
	\footnotesize
	\begin{tabular}{@{}l l l l@{}}
		\toprule
		Scenario Acronym & \makecell{Single-node /                                    \\ Multi-node} & \makecell{Network \\ Medium} & \makecell{Real-world \\ Examples} \\ \midrule
		CPU Offload      & Single-node             & N.A.     & \makecell{Single-node \\ CPU Offloading} \\ \hline
		Central Storage  & Single-node             & Ethernet & \makecell{Centralized \\ Storage Server} \\ \hline
		PD               & Single-node             & NVLink   & \makecell{PD          \\ Disaggregation} \\
		\bottomrule
	\end{tabular}
	\caption{Evaluation scenarios setup. }
	\label{tab:eval_setup}
\end{table}

We evaluate \name under three different scenarios, as shown in Table~\ref{tab:eval_setup}.
The three scenarios are representative setups that are commonly used by the users of \name.


\mypara{Models} We compare \name against baseline solutions on popular open source models adopted by industry: \texttt{meta-llama/Llama-3.1-8B-Instruct}, 
\texttt{Sao10K-L3-8B}, \texttt{meta-llama/Llama-3.1-70B-Instruct}, \texttt{Qwen/Qwen2.5-Coder-32B-Instruct}, \texttt{Qwen/Qwen3-Coder-480B-A35B-Instruct-FP8}, \texttt{Qwen/Qwen2.5-72B-Instruct}.

\mypara{Datasets} \name is evaluated on several datasets, including emulated multi-round question answering, long context question answering from LongBench~\cite{bai2024longbenchbilingualmultitaskbenchmark}, and random dataset from vLLM official benchmarking script~\cite{vllm}.

\mypara{Hardware} For single-node evaluation, we run \name on an 8$\times$H100 server provided by GMI Cloud~\cite{gmicloud2025}.
Because different models require varying numbers of GPUs to be served, we allocate the minimum number of H100 GPUs necessary to successfully start each model in our evaluation.
For multi-node evaluation, we use the same number of GPUs as in the single-node setup and configure a centralized remote storage backend that leverages CPU memory for KV cache storage.
For PD disaggregation, the prefiller and decoder instances are both set up with the number of GPUs as in single-node evaluation, and the prefiller and decoder instances are connected with NVLink.

\mypara{Metrics} For each experiment, we show both time-to-first-token (TTFT), which is the prefill delay, and inter-token-latency (ITL), which is the average delay between the generation of two consecutive output tokens.
For component-wise analysis which breaks down the delay for CPU offloading or PD disaggregation, we report the delay for each component separately.

\mypara{Baselines} We compare \name \texttt{v0.3.6} with several baselines, including:
\begin{packeditemize}
    \item Basic vLLM: vLLM \texttt{v0.10.2} which enables prefix caching by default, but only keeps KV cache inside GPU memory, so only a small portion of it can be kept; 
    \item Basic vLLM CPU Offloading: vLLM \texttt{v0.11.0} with its own implementation of CPU offloading;
    \item Commercial offerings \#1 and \#2:  provides dedicated endpoint service that reserves GPUs for users to run a user-defined model. We ran these baselines accessed on September 10th. 
\end{packeditemize}


\subsection{Single-node CPU Offloading}

We first evaluate \name on the CPU Offload scenario as in Table~\ref{tab:eval_setup}.
In this experiment, we use multi-round Q\&A workloads that emulate a typical chatbot-based document analysis scenario. By default, each LLM query contains 10K tokens, consisting of a document (roughly a 12-page PDF) used as context and a unique short question. Llama-3.1-8B-Instruct model takes 20K tokens as input, since smaller models are generally better can handle more and longer queries.
The LLM output is a short answer of 100 tokens at max.
The chat session begins with 40 users, and additional users join according to a specified arrival rate (QPS).
We set the maximum CPU memory \name can offload KV cache to 500~GB.

As shown in Figure~\ref{fig:single-node}, \name consistently outperforms all baselines in both TTFT and ITL. For instance, under low QPS (e.g., QPS = 1), \name has 1.9 to 8.1$\times$ smaller TTFT. 
\name achieves 2.3--14$\times$ higher query processing rate (\ie throughput), at the same TTFT, than the strongest baseline across five evaluated models. In terms of ITL, \name also outperforms the baselines, as they incur a long delay before generating the first token, which in turn causes subsequent token generation to be queued.
Specifically, compared to the best baseline, \name has 7\% to 92\% smaller ITL, at QPS=1. 
For Qwen3-Coder-480B, commercial options \#1 and \#2 do not provide support for hosting the model.

\mypara{Understanding \name's gains}
\name outperforms baselines for several reasons. Compared with basic vLLM, which caches KV data only in GPU memory, \name leverages CPU offloading. Since CPU memory can hold far more KV cache than GPU memory, \name achieves significantly higher cache hit ratios. 
Compared to basic vLLM CPU offloading, which employs per-layer and per-16-token transmission and is unable to fully utilize the loading bandwidth, \name has more efficient implementation of the transmission module as it loads KV cache at chunk level with high-performance data loading CUDA kernels. 
Our comparison with closed-source commercial alternatives is conducted in a black-box manner since their internal implementations are not publicly available.
From the end-to-end results, we hypothesize that Commercial Option \#1 lacks a KV cache offloading mechanism to secondary storage.
In contrast, Commercial Option \#2 likely supports KV cache offloading to secondary storage, yet its performance is still worse than \name. 



\subsection{Real-trace Driven Evaluation}
\begin{figure}[t]
    \centering
    \includegraphics[]{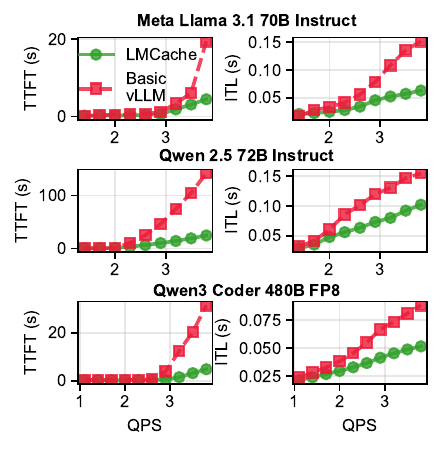}
    \tightcaption{Comparing \name and basic vLLM on three different models based on real trace drawn from company F's input and output distributions. \name has at least 4.4-6.6$\times$ smaller TTFT, and 34-58\% smaller ITL, at high QPS. }
    \label{fig:real-trace}
\end{figure}

\begin{figure}[t]
    \centering
    \includegraphics[]{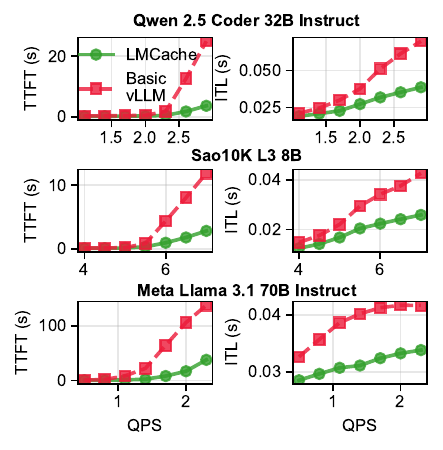}
    \tightcaption{Upper and middle figures: comparing \name and basic vLLM on Qwen 2.5 Coder 32B Instruct and Sao10K L3 8B models based on real trace drawn from company F. 
    Lower figure: comparing \name and basic vLLM on Llama 3.1 70B Instruct on company drawn from company G's input/output data distribution. 
    \name achieves 3.7-6.8$\times$ smaller TTFT and 19--44\% smaller ITL. }
    \label{fig:real-trace-2}
\end{figure}

We evaluate \name on a real trace drawn from company F and company G's distribution of input and output tokens. 
Since we do not have access to company $F$’s proprietary models, we run the trace using five different models. To make the experiment tractable, we stretch the original trace which lasts for several days so that the workload we run completes within one hour.
\name is set to use 500~GB of CPU DRAM at maximum, and we compare it with latest basic vLLM with GPU prefix caching. 

As shown in Figure~\ref{fig:real-trace} and Figure~\ref{fig:real-trace-2}, \name consistently outperforms basic vLLM on the real trace on different QPS across the five models. 
Specifically, \name reduces TTFT by at least 3.7--6.8$\times$, and reduces ITL by at least 19-58\%, across five models at high QPS. 

\subsection{Centralized Storage Server}
\begin{figure*}[t]
	\centering
	\includegraphics[]{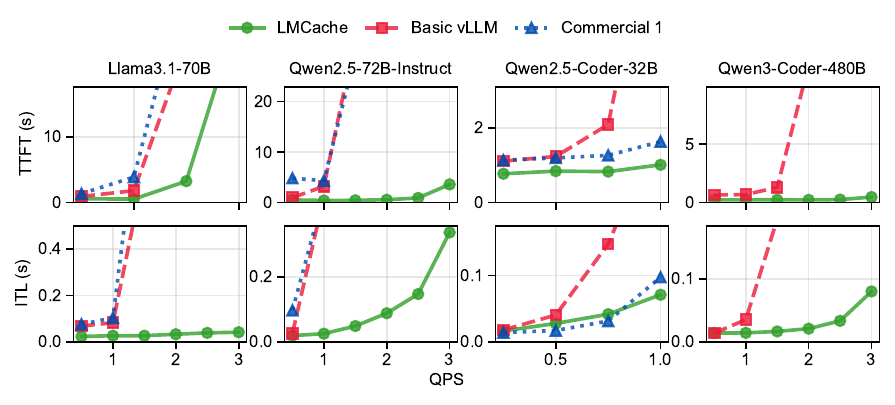}
	\tightcaption{Compared to basic vLLM, \name with remote backend offloading has 1.3 to 3$\times$ improvement in inference throughput, under the same TTFT. }
	\label{fig:remote-backend}
\end{figure*}

Next, we run \name for KV cache sharing through a centralized remote server, that is connected to the GPU instance with a bandwidth of 15 Gbps, following the setup of central storage in Table~\ref{tab:eval_setup}. For this experiment, we evaluate using the TriviaQA dataset from LongBench~\cite{bai2024longbenchbilingualmultitaskbenchmark}, a widely adopted benchmark for long-context evaluation. We follow the official vLLM benchmarking scripts~\cite{vllm}, which generate inference queries according to a Poisson distribution at a specified QPS.

As shown in Figure~\ref{fig:remote-backend}, \name consistently outperforms all baselines across different QPS levels, providing 1.3--3$\times$ improvement in inference throughput. The improvement comes from the fact that the remote backend can store far more KV cache than CPU memory, thereby achieving higher cache hit ratios.

We note, however, that loading KV cache from the remote backend introduces greater latency than loading from CPU memory, since the remote backend has a much lower bandwidth. As a result, the loading delay may even surpass the prefill delay, particularly when the input context is short or model is small, as prefilling is too fast in such cases—a scenario that we will demonstrate later in~\ref{subsec:sens_study}.
Thus, adaptive decisions between KV cache loading and prefilling need to be made when KV cache resides in a remote storage server. 




\subsection{PD disaggregation}
\label{subsec:pd-eval}
\begin{figure*}[t]
	\centering
	\includegraphics[]{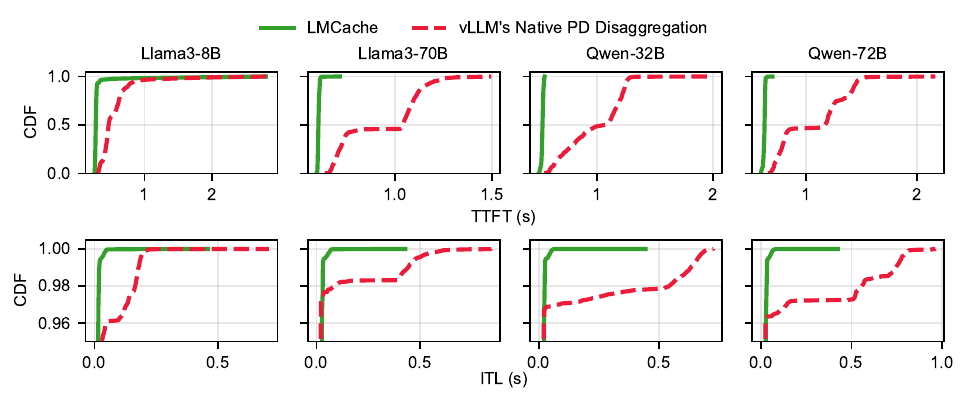}
	\tightcaption{Compared to vLLM's native PD disaggregation, \name's PD disaggregation has significantly lower tail latency, and achieving 1.5--1.8$\times$ lower mean TTFT, and 1.1 to 1.7$\times$ lower mean ITL. }
	\label{fig:pd-cdf}
\end{figure*}

\begin{figure}[t]
	\centering
	\includegraphics[]{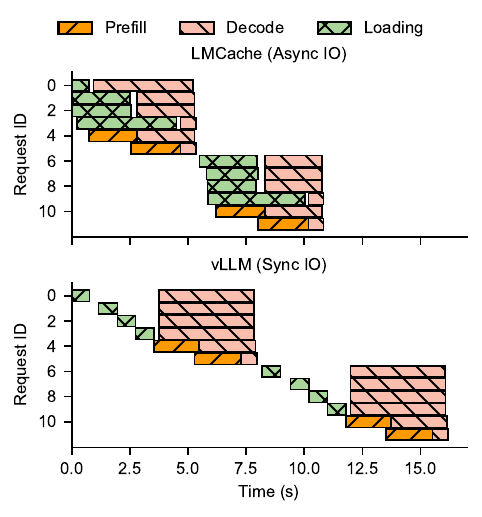}
	\tightcaption{With request asynchronization, \name overlaps KV cache loading and inference computation (either prefill or decode). 
    }
	\label{fig:async}
\end{figure}

\begin{figure}[t]
	\centering
	\includegraphics[]{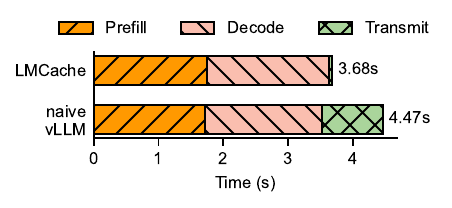}
	\tightcaption{Compared to vLLM's native PD disaggregation, \name achieves much smaller transmission latency, thus reducing end-to-end delay. }
	\label{fig:pd-breakdown}
\end{figure}


In this experiment, we evaluate the performance in a PD disaggregation setting.
Here, we compare \name with vLLM’s native PD disaggregation with the official benchmarking script for random input and output workload. We use 8K tokens input and 200 tokens output.
As shown in Figure~\ref{fig:pd-cdf}, it presents the 95th percentile TTFT for both \name and vLLM’s native PD disaggregation, showing that \name achieves significantly better tail latency.
In terms of mean TTFT, \name also greatly outperforms vLLM native PD disaggregation.
Specifically, \name reduces mean TTFT by 1.53--1.84$\times$, and reduces mean ITL by 1.12--1.66$\times$, across the four models.

The performance gains of \name over the baseline stem from its more efficient design for PD disaggregation. Specifically, \name copies each chunk of the KV cache (generated during chunked prefill) to a buffer in the GPU memory of the prefiller instance, and then transfers it to the corresponding buffer on the decoder instance. Once received, the KV cache is copied into the paged memory of the decoder instance.

In contrast, vLLM’s native PD disaggregation sends the paged KV cache generated by the prefiller directly to the decoder, using NIXL’s memory copy function. This function takes as input the memory addresses of the KV cache pages on the prefiller side and copies them to the destination addresses on the decoder side. However, when the paged memory for the KV cache is scattered across the prefiller’s GPU memory, the transfer is performed in a page by page manner, which leads to bandwidth underutilization, as discussed in \S\ref{sec:design}.



\subsection{Component-wise Evaluation}

To further understand the gain brought by \name, we also perform component-wise analysis to break down the delay of each component in the end-to-end system.

\mypara{PD disaggregation} Figure~\ref{fig:pd-breakdown} shows the latency breakdown of LLM inference, including both prefill and decode computation, as well as the transmission of KV cache between prefiller and decoder instances. The prefill and decode computation times are the same for \name and vLLM's native PD disaggregation.
However, as discussed in \S\ref{subsec:pd-eval}, vLLM's native design transmits KV cache at a much finer granularity, which results in bandwidth underutilization.
In contrast, \name employs a more efficient KV cache transfer mechanism, enabling significantly faster transmission and thereby reducing the overall end-to-end delay in PD disaggregation.

\begin{table}[]
	\begin{tabular}{@{}ll@{}}
		\toprule
		Method                       & Achieved Bandwidth \\ \midrule
		LMCache                      & 400 Gbps           \\
		vLLM's Native CPU Offloading & 88 Gbps            \\ \bottomrule
	\end{tabular}
	\tightcaption{\name achieves much higher loading bandwidth when loading KV cache from CPU memory, compared to vLLM's native CPU offloading. }
    \label{tab:cpu_speed}
\end{table}
\mypara{CPU offloading} In Table~\ref{tab:cpu_speed}, we perform an ablation study to test the achieved loading bandwidth from CPU for \name and vLLM's native CPU offloading.
The reason \name achieves higher transfer bandwidth than vLLM’s native CPU offloading is due to the transfer granularity.
While native CPU offloading performs data movement page by page, \name transfers data chunk by chunk. Each transfer operation triggers a CUDA memory copy, which involves preparing metadata beforehand and sending a completion signal afterward.
These per-transfer operations add overhead to every memory copy kernel.
By transferring larger chunks of data per copy, \name reduces the overall overhead, resulting in a much higher effective bandwidth.

\mypara{Asynchronous Compute} We also show the benefit of \name's asynchronous compute in terms of reducing end-to-end delay.
Figure~\ref{fig:async} shows the timeline of queries loading and inference computation.
The figure is drawn from the middle of a longer run for illustration purpose.
As shown in the figure, without query asynchronization, prefill/decode computation and loading happen sequentially.
With query asynchronization, the prefill/decode computation can overlap with KV cache loading, which reduces the end-to-end delay by 1.46$\times$.

\subsection{Sensitivity Study}
\label{subsec:sens_study}

We also perform several sensitivity evaluation to see how \name's delay changes under different context lengths and different types of remote backends.
\begin{figure}[t]
	\centering
	\includegraphics[]{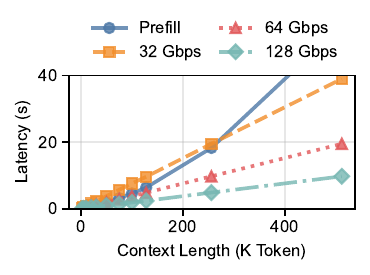}
	\tightcaption{At network bandwidth of 32Gbps, \name's KV cache offloading only outperforms basic vLLM's prefill when input length is more than 256K tokens. 
    At network bandwidth of 64 or 128Gbps, \name's KV cache offloading is better than prefill across all input lengths. }
	\label{fig:diff-lengths}
\end{figure}


\mypara{Impact of context lengths}
Figure~\ref{fig:diff-lengths} shows the prefill delay on B200 machines and the loading delay under different network bandwidths.
When the network bandwidth is low (\ie 32 Gbps), \name's KV cache loading outperforms naive prefilling only when the input context length exceeds 256K tokens.
In contrast, when the bandwidth is higher (\ie 64 or 128 Gbps), \name's loading consistently achieves lower delay than naive prefilling across all context lengths.
These results suggest that \name's KV cache loading should be adaptive: under low bandwidth, loading should be enabled only when the context length surpasses the crossover point where loading becomes faster than prefilling.


\subsection{SGLang Results}
\begin{figure}[t]
	\centering
	\includegraphics[]{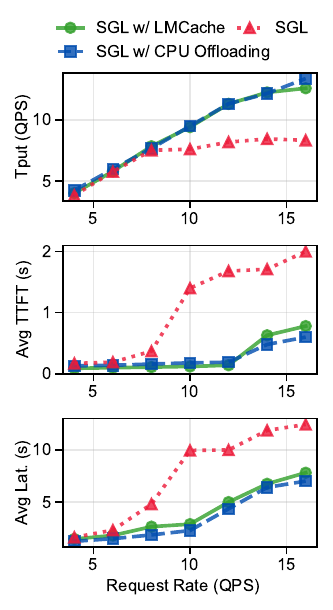}
	\tightcaption{On \texttt{Qwen3-32B} model, \name's CPU offloading achieves comparable performance as SGLang's native CPU offloading. }
	\label{fig:sglang}
\end{figure}

Although our primary evaluation uses vLLM, we also evaluate \name integrated with SGLang.
Figure~\ref{fig:sglang} reports results for Qwen3-32B served on two H100 GPUs (TP=2) with \name’s CPU offloading enabled.
Compared to SGLang without CPU offloading, \name achieves higher throughput and lower mean TTFT and mean end-to-end latency.
Compared to SGLang’s native CPU offloading, \name achieves comparable performance.
These results confirm that \name is also effective on another inference engine.
Although SGLang’s native CPU offloading achieves performance comparable to \name’s CPU offloading on SGLang, it lacks a distributed storage backend capable of efficiently offloading data across a hierarchical set of storage devices, such as local disks and remote CPU/disk resources.

\section{Real-World Lessons and Experience}
\label{sec:deployment}
\mypara{Loading from remote storage is faster than prefill} Traditionally, it was believed that loading KV cache from remote storage primarily improves cache hit rates and reduces storage costs by leveraging cheaper storage devices, but at the cost of increased inference latency, since loading data from remote devices was thought to be slower than performing a full prefill.
This assumption was largely due to the historically low throughput of remote object stores such as Amazon S3, which offered loading speeds as low as 100 MBps.
Recently, however, there has been a major boost in remote storage performance, for example, with Amazon S3 Express, the throughput has increased from 100 MBps to nearly 1 GBps.
Our users, such as Company C, have adopted \name to load KV caches from their own remote object store, achieving 22–32\% lower TTFT compared to full prefill.
This insight suggests that remote backends can simultaneously improve cache hit ratios and reduce TTFT.

\mypara{Context truncation lowers prefix cache hit rates}
Many industry users employ a sliding-window mechanism to handle long-context inputs constrained by limited model context windows or GPU memory.
For instance, when input tokens exceed the context window limit, some companies truncate the input to keep only the most recent tokens.
However, this approach significantly reduces prefix cache hit ratios, since truncated inputs no longer match the prefixes of previously cached contexts.
In practice, using real traces from Company F, we observe that prefix cache hit ratios drop from roughly 85\% to 45\% when truncating input contexts to keep only the latest tokens.
Other studies have also discussed this phenomenon, emphasizing that dynamically adding or removing context tokens should be avoided, as it invalidates prefix KV cache reuse~\cite{manus_blog}.

\mypara{Containerized code is preferred} With the growing scale of LLM inference, most production environments now rely on Kubernetes to manage GPU clusters.
Consequently, deploying inference engines (\eg vLLM or SGLang) and \name through containerized environments, typically via Docker images, has become the standard practice among industry users.
Interestingly, many users rely solely on the official Docker images without going deep into \name’s source code.

\mypara{Unexpected high cache hit rate in production systems} Our customers did not expect such a high prefix cache hit ratio, such as 50\% hit rate for company G in their production environments, until they deploy \name inside their systems. 
Previously, people thought KV caches could only be reused for fixed system prompts.
However, modern applications increasingly exhibit "dynamically reusable contexts", such as conversation histories in coding assistants, chat applications, and retrieval-augmented generation (RAG) pipelines.
These emerging patterns have significantly increased overall cache hit ratios in real-world deployments.

\mypara{Industry vs. academia users} \name was designed as a unified prototype framework in May 2024, where we can put our research works into it to gain more impact. 
However, we found that at that time, industry wants a highly efficient KV cache offloading solution as the size of KV cache and concurrent users keep growing. 
Then, our attention has been moved towards improving the performance, stability, and compatibility. 
Since most companies are less concerned with customizing attention algorithms, we deprioritized designing flexible APIs for integrating specialized attention mechanisms, such as selective token dropping.
This makes \name less popular in academia, since the academia users often focuses on modifying attention mechanisms for their research prototypes. 
As next step, \name is going to design more flexible APIs such that it is easy to use by both industry and academia.

\mypara{Flexibility vs. performance for programming language} Python has always been the de facto language in ML. However, the current industry focus is gradually shifting towards higher efficiency instead of broad compatibility. 
Specifically, many companies are rewriting ML libraries in high-performance languages such as Rust or C++, or carefully optimizing Python-based systems to hide its runtime overhead while preserving flexibility.
Although some alternatives have been writing their ML libraries in Rust, we continue to use Python with carefully designed optimizations.
This approach allows us to evolve faster, with more community contributions, and still maintain performance on par with  alternatives.

\mypara{\name now a community effort} One key reason \name rapidly evolved from a research prototype to a widely adopted industry framework is the active involvement of community contributors.
About a year ago, \name supported only local CPU, local disk, and Redis backends integrated with vLLM on NVIDIA GPUs.
Today, it supports eight more storage backends (NFS, WEKA, GPU-Direct Storage, Mooncake Store, NIXL, S3, InfiniStore, and Valkey) across four processor types (NVIDIA, AMD, Ascend, and TPU), and two inference engines (vLLM and SGLang). 
All of these contributions were made by industry partners who actively upstreamed their code to stay aligned with ongoing development and avoid divergence from the latest \name updates.


\section{Conclusion and Outlook}

This paper presented \name, the first open-source and most widely adopted production-ready KV caching layer for enterprise-scale LLM inference. 
By treating KV cache as a first-class data structure rather than an internal byproduct of inference, \name transforms LLM engines from isolated token processors into a distributed ecosystem of compute and storage. Evaluation across diverse workloads and models demonstrates that \name consistently delivers significant throughput improvement and latency reduction compared to both open-source baselines and commercial inference APIs. 
Beyond performance, \name has already seen rapid adoption in production environments, where enterprises leverage its CPU offloading, hierarchical storage, and PD disaggregation capabilities to keep low latency and reduce cost in trillion-token–scale deployments. 
Real-world deployments have also revealed new opportunities, such as KV cache reuse in recommendation systems and lossy compression in open-ended chatbots, underscoring the versatility of \name across application domains.

Looking ahead, \name points to a broader shift: {\bf AI-native data such as KV caches will increasingly serve as the substrate for scaling LLM inference and agentic workloads}. 
By establishing KV cache as a standardized storage and communication medium, \name lays the foundation for future systems that treat inference not as isolated sessions but as a persistent, cache-aware computation fabric. 
We hope that the design, optimizations, and deployment lessons presented in this paper will inform the next generation of LLM infrastructure, where AI-native data, such as KV caches, is not merely an optimization but a core primitive for efficient, reliable, and scalable inference.

The source code of \name is at: \url{https://github.com/LMCache/LMCache}.

\tightsection{Acknowledgement}

We would like to thank the \name community for their invaluable support and contributions, including Baolong Mao and Chunxiao Zheng for managing remote connectors, Martin Hickey for GitHub Infrastructure, Huaizheng Zhang, Siddhant Ray, Zhuohan Gu and Hanchen Li for writing and maintaining documentation, Qizheng Zhang and Hussain Mohammad for insightful feedback. 
We also thank GMI cloud for providing us GPU clusters to run the experiments.


\newpage
\bibliographystyle{plainnat}
\bibliography{citations}

\end{document}